# SDRCNN: A single-scale dense residual connected convolutional neural network for pansharpening

Yuan Fang, Yuanzhi Cai, *Member, IEEE*, and Lei Fan, *Member, IEEE*

*Abstract*—**Pansharpening is a process of fusing a high spatial resolution panchromatic image and a low spatial resolution multispectral image to create a high-resolution multispectral image. A novel single-branch, single-scale lightweight convolutional neural network, named SDRCNN, is developed in this study. By using a novel dense residual connected structure and convolution block, SDRCNN achieved a better trade-off between accuracy and efficiency. The performance of SDRCNN was tested using four datasets from the WorldView-3, WorldView-2 and QuickBird satellites. The compared methods include eight traditional methods (i.e., GS, GSA, PRACS, BDSD, SFIM, GLP-CBD, CDIF and LRTCFPan) and five lightweight deep learning methods (i.e., PNN, PanNet, BayesianNet, DMDNet and FusionNet). Based on a visual inspection of the pansharpened images created and the associated absolute residual maps, SDRCNN exhibited least spatial detail blurring and spectral distortion, amongst all the methods considered. The values of the quantitative evaluation metrics were closest to their ideal values when SDRCNN was used. The processing time of SDRCNN was also the shortest among all methods tested. Finally, the effectiveness of each component in the SDRCNN was demonstrated in ablation experiments. All of these confirmed the superiority of SDRCNN.**

*Index Terms*— **Pansharpening, convolutional neural network, multispectral image, fusion, deep learning, resolution.**

## I. INTRODUCTION

WITH the rapid development of earth observation satellites, remotely sensed images have widely been used for various applications such as object detection and semantic segmentation [1]–[4]. However, due to the physical constraints of the existing single sensors, a trade-off between the spectral resolution and the spatial resolution of an image to be acquired needs to be considered [5], [6]. Therefore, remote sensing satellites typically carry two imaging sensors to capture multispectral (MS) images and panchromatic (PAN) images, respectively. MS images consist of multiple bands at the cost of a relatively low spatial resolution, while PAN images contain finer spatial details in a single image band. To combine their advantages, pansharpening is a typical technique used to fuse an MS image and a PAN image to form a high resolution multispectral (HRMS) image that shares the same spatial resolution of the PAN image and the same spectral resolution of the MS image. An example of this process is illustrated in Fig. 1. Pansharpening has become a pre-processing step of image enhancement in many remote sensing tasks, such as object detection [7], [8], anomaly detection [9], [10] and agricultural management [11], [12].

Over the last few decades, various methods have been proposed to achieve pansharpening. They can be divided into four categories, including component substitution (CS) methods, multi-resolution analysis (MRA) methods, variational optimization (VO) techniques and deep learning (DL) approaches. CS-based methods have widely been used due to their simple principle and easy implementation. They project an interpolated MS image into a transformed domain to separate the spectral and the spatial information. The separated spatial components are then replaced by a PAN image, followed by an inverse transformation to generate the HRMS image. Representative CS-based methods include intensity–hue–saturation (IHS) [13], principal component analysis (PCA) [14], Gram–Schmidt adaptive (GSA) technique [15], partial replacement adaptive CS (PRACS) [16] and band-related spatial detail (BDSD) scheme [17]. These CS-based methods are effective in increasing spatial resolution, but incur spectral distortion [18]. MRA-based methods inject spatial details extracted from PAN images by spatial filtering into interpolated MS images. Spatial details can be extracted by different decomposition methods. Compared to CS-based methods, this class of methods preserves spectral information well. However, artifacts are prone to occur due to aliasing effects, resulting in spatial distortion [6]. MRA-based methods include wavelet transform [19], smoothing filter-based intensity modulation (SFIM) [20], additive wavelet luminance proportion (AWLP) [21], and the modulation transfer function generalized Laplacian pyramid with full resolution regression-based injection model (GLP-Reg) [22]. VO-based methods mainly need to construct and optimize an energy function. After the P+XS [23] method was proposed, such methods have received more attention and research efforts. They can well model the relationships between PAN, MS and HRMS images and produce a high-quality image fusion. However, compared to CS-based and MRA-based pansharpening methods, VO-based methods are complex in operation and computationally expensive, which limit their wide applications [24]–[26].

In recent years, the successful application of deep learning

This research was funded by Xi'an Jiaotong-Liverpool University Research Enhancement Fund, grant number REF-21-01-003, Xi'an Jiaotong-Liverpool University Postgraduate Research Scholarship, grant number PGRS2006010 *(Corresponding author: Lei Fan).*

Yuan Fang, Yuanzhi Cai, and Lei Fan are with Department of Civil Engineering, Design School, Xi'an Jiaotong-Liverpool University, Suzhou, 215000, China, and also with the School of Engineering, University of Liverpool, L69 3BX Liverpool, U.K. (e-mail: yuan.fang16@student.xjtlu.edu.cn; yuanzhi.cai19@student.xjtlu.edu.cn; lei.fan@xjtlu.edu.cn).

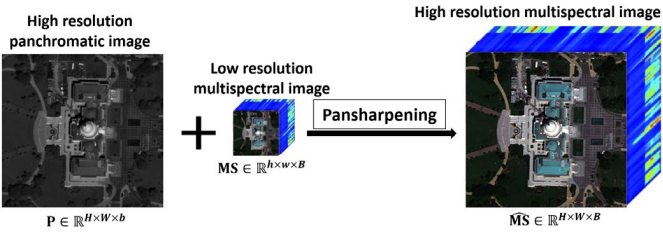

**Fig. 1.** An example of pansharpening: $H \times W$ and $h \times w$ represent the image size ($h < H$ and $w < W$), $b$ and $B$ represent the number of image bands ($1 = b < B$).

in computer vision has attracted a lot of attention [27]–[32]. In particular, convolutional neural networks (CNNs) stand out for powerful non-linear modelling capabilities and have made significant progress in pansharpening [33]–[35]. As a pioneering attempt, Masi et al [33] proposed a simple three-layer pansharpening neural network (PNN) that was based on a modification of the super-resolution convolutional neural network (SRCNN) [36] and led to promising results. Based on PNN, Wei et al. proposed a deeper residual pansharpening network (DRPNN) [37]. Yuan et al. introduced multilevel feature extraction and proposed a multi-scale multi-depth CNN (MSDCNN) [38]. Shao et al. proposed a two-branch network called RSIFNN [39]. It was noticed that the majority of deep learning improve the pansharpening accuracy by utilising complex network structure designs and/or diverse computational units [40], which are often achieved by using multi-scale structures [41]–[44], multi-branch structures [45]–[48], deeper network [49], generative adversarial networks [50]–[53], attention mechanism [54]–[56], and transformer modules [57]–[59]. However, these would make CNNs far larger in size (i.e., number of parameters) than PNN, and consequently lead to reduced computational efficiency [49], [60], [61].

The efficiency of the panchromatic sharpening methods is important for practical applications, especially considering the huge volume of data captured by satellite. Therefore, it is of interest to develop a lightweight network for higher pansharpening accuracy. To this end, a single-branch, single-scale, lightweight convolutional network named SDRCNN is developed in this study. In SDRCNN, novel convolutional block and dense residual connected structures are designed to recover spatial details from coarse to fine. The effectiveness and the efficiency of SDRCNN are tested using multispectral images acquired by three satellites, including WorldView-2, WorldView-3 and QuickBird. The major contributions of this paper are:

1) Development of a lightweight pansharpening network (i.e., SDRCNN) to achieve a better trade-off between accuracy and efficiency.
2) A comprehensive fair (i.e., with similar numbers of parameters) evaluation of the performance of various lightweight network structures and the convolutional blocks.

This paper is organised as follows. Section II describes the methodology. Experimental results and the associated discussions are presented in Sections III and IV, respectively.

Finally, conclusions are drawn in Section V.

## II. METHODOLOGY

### A. Overall Architecture of SDRCNN

As illustrated in Fig. 2, the proposed SDRCNN takes a PAN image and an upsampled LRMS image as the inputs. Different levels of structural details are extracted through a Stem Block and three Residual Blocks, which are arranged in a concatenated manner. Using the proposed dense residual connection mechanism, 52 feature maps are obtained at each of the three different network depths (i.e., after the three Residual Blocks). In addition to the Stem Block, reaching these three network depths requires passing one, two and three identical residual blocks, respectively. The 156 feature maps are concatenated and passed through a layer of $1 \times 1$ convolution operations to enable the transfer of different levels of feature information, generating a data cube with the same number of channels as the input MS image. The output data cube is then directly summed with the input upsampled LRMS to produce the final HRMS image. The L1 loss is adopted in the network as the loss function in this study. An elaboration of proposed SDRCNN is presented in Sections II.B, II.C and II.D.

### B. Pre-processing

The existing CNN-based pansharpening methods usually rely on large-scale training datasets to learn the nonlinear mapping between the input PAN and MS images and the ground-truth HRMS image. However, this process treats pansharpening as an image regression problem in a black-box learning process. Ideally, pansharpening is to inject useful spatial details from the input PAN image into the input MS image. In order to better preserve the spectral information, our network directly maps an input upsampled MS image to the output of the network, which achieves lossless propagation of the spectral information on the one hand, and makes the network more focused on the extraction of spatial features on the other, thus significantly reducing the difficulty of network learning. The ablation experiment results in Section III-E-3) show that adding spectral mapping can improve the spectral accuracy of the fused image.

### C. Dense Residual Connection

To further improve the learning capability and efficiency of feature extraction in SDRCNN, we propose a dense residual connection strategy, as shown in Fig. 2. As the key idea of this strategy, three different levels of feature maps are integrated and added to the input upsampled LRMS to obtain HRMS images with more spatial details. In other words, features output from Residual Blocks of different depths are stacked with the feature information output from all previous blocks and fed into the next Residual Block, thus making fuller use of the residual information at different levels to learn features from coarse to fine to enhance the network's ability on reconstructing the spatial information of the fused image. The inputs to the Stem Block and each of the three residual blocks in SDRCNN can be expressed as the following formulae:

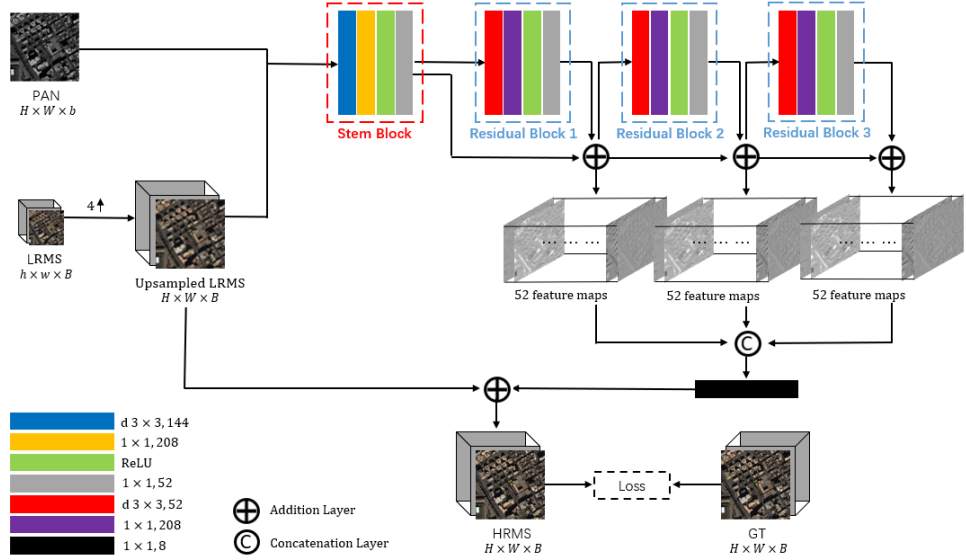

**Fig. 2.** Overall architecture of proposed SDRCNN.

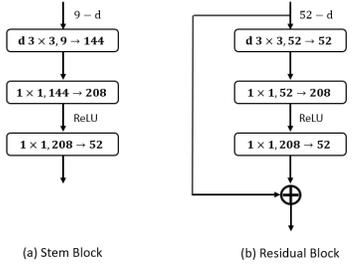

**Fig. 3.** Illustration of the convolution blocks used in SDRCNN: (a) Stem Block, and (b) Residual Block.

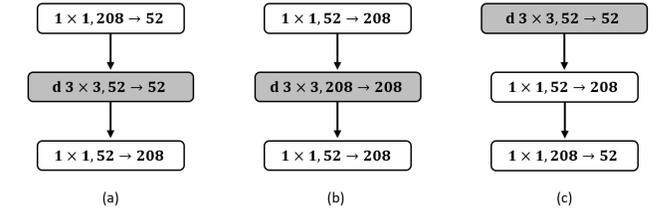

**Fig. 4.** Process of modifying the block configuration and the associated specifications: (a) a bottleneck block; (b) an inverted bottleneck block; (c) the adopted block with the position of the spatial depthwise convolutional layer being moved to top.

$$I_S = PAN + LRMS \uparrow_4 \qquad (1)$$

$$I_R^1 = F_S \qquad (2)$$

$$I_R^2 = F_S \oplus F_R^1 \qquad (3)$$

$$I_R^3 = (F_S \oplus F_R^1) \oplus F_R^2 \qquad (4)$$

where $I_S$ and $F_S$ are the input and output feature maps of the Stem Block, respectively; $I_R^i$ and $F_R^i$ are the input and output feature maps of the $i$ th Residual Block, respectively; $\uparrow_4$ represents the operation of up-sampling by a factor of 4; + indicates a series connection; $\oplus$ indicates the operation of adding the values of each element.

Using this dense residual connection mechanism, the Addition Layer following each Residual Block numerically sums the residuals obtained from all blocks before that network depth and forms a set of feature maps (i.e., 52 maps). The output features from a deeper layer would include more spatial information. In total, 156 feature maps from the three levels are concatenated to propagate and integrate the information at different levels through a $1 \times 1$ convolution operation. The generated residual information is summed with the input upsampled LRMS to produce a fused HRMS image. This dense residual connection mechanism allows the proposed single-branch, single-scale SDRCNN to learn diverse (i.e., different levels) features similar to that of complex (e.g., multi-branch, multi-scale) networks.

### D. Structure of Functional Blocks

Two major types of convolution blocks are used in this study, i.e., Stem Block and Residual Block. Their detailed structures are shown in Fig. 3. The Stem Block is used to integrate the input images of the network into a data cube with pre-determined width to obtain the number of channels appropriate to the Residual Block. The Residual Blocks are used for feature extraction and processing.

Inspired by the well-known network ResNet [62], the same residual connections are used in our Residual Blocks, which can solve the problem of exploding or disappearing gradients. In order to reduce the number of parameters to be learned for convolutional computation, all the $3 \times 3$ convolutional layers in the network use depthwise separable convolution [63], thus improving the computational efficiency of SDRCNN. Unlike the usual bottleneck configuration (Fig. 4(a)), the inverted bottleneck structure (Fig. 4(b)) was employed to avoid the loss from compressed dimensions during the information transformations between different dimensional feature spaces. However, because the inverted bottleneck amplifies the intermediate convolutional layers, a direct replacement will lead to an increase in the number of parameters. Therefore, the depthwise separable convolutional layer is moved to the top of

| **Traditional methods** |
| --- |
| GS: Gram-Schmidt transformation [73] |
| GSA: Gram–Schmidt adaptive [15] |
| PRACS: Partial replacement adaptive CS approach [16] |
| BDSD: Band-related spatial detail scheme [17] |
| SFIM: Smoothing filter intensity modulation [20] |
| GLP-CBD: GLP with MTF-matched filter and regression-based injection model [74], [75] |
| CDIF: Context-Aware Details injection fidelity with adaptive coefficients estimation for variational pansharpening [76] |
| LRTCFPan: Low-Rank tensor completion-based framework for pansharpening [77] |
| **DL-based methods** |
| PNN: CNN for pansharpening [33] |
| PanNet: A deep network architecture for pan-sharpening [65] |
| BayesianNet: Bayesian pansharpening with multiorder Gradient-based deep network constraints [66] |
| DMDNet: Deep multiscale detail networks for multiband spectral image sharpening [67] |
| FusionNet: Deep CNN inspired by traditional CS and MRA methods [60] |

a block (Fig. 4 (c)) to reduce the computational effort. This block configuration (Fig. 4(c)) is adopted for the Stem Block and the Residual Blocks in SDRCNN.

In addition, the two convolution blocks (i.e., the Stem Block and the Residual Block) used in this study removed the commonly used batch normalization (BN) layers. For most cases, using BN can accelerate the training and lower the sensitivity of the network to initialization. However, our trial experiments indicated that using BN did not consistently improve the pansharpening accuracy. Simultaneously, due to the additional computation and memory usage of executing BN, the BN layers inside the ResNet Block are discarded in our approach. Finally, in each block, only one ReLU activation function layer is set between two $1 \times 1$ layers.

### E. Datasets

Four datasets from WorldView-2, WorldView-3 and QuickBird (three different sensors working in the visible and near-infrared spectral ranges) were used to test our network SDRCNN in this article. WorldView-2 and WorldView-3 provide MS images with eight bands (coastal, blue, green, yellow, red, red edge, near-infrared 1 and near-infrared 2) and single-band PAN images. The spatial resolutions of these MS and PAN images are approximately 1.8 m and 0.5 m respectively for WorldView-2, 1.2 m and 0.3 m respectively for WorldView-3. QuickBird provides MS images with four bands (red, green, blue, and near-infrared) and single-band PAN images. The spatial resolutions are approximately 2.44 m for MS images and 0.61 m for PAN images.

### F. Implementation Simulation

#### 1) Dataset Simulation

Since there is no ground truth (GT) image as a reference, the Wald's protocol [64] was adopted. According to this protocol, the original MS images were used as the GT images. In addition, the original MS and PAN images were simultaneously blurred and downsampled to produce modified images that were used as input images. Taking the two WorldView-3 datasets (i.e., Tripoli and Rio) as an example, each dataset was simulated with 12580 samples (also called patch pairs), in which each sample includes PAN (256×256 in size), LRMS (64×64×8 in size) and GT (256×256×8 in size) patches. These 12580 samples were randomly split into 70%, 20% and 10% for training, validation and test, respectively. Furthermore, another 8-band dataset (i.e., WorldView-2) and a 4-band dataset (i.e., QuickBird) were also used for performance evaluations in this study.

#### 2) Benchmark

To check the performance of our network SDRCNN, several representative pansharpening methods of various categories were implemented for comparisons, as summarized in Table I. The traditional methods were considered because they are still widely used in applications. In terms of deep learning-based methods, PNN, PanNet [65] and three state-of-the-art lightweight networks with the number of parameters being consistent with PNN were considered. For fair comparisons, PanNet, BayesianNet [66], DMDNet [67], FusionNet [60] and SDRCNN were adjusted to have almost the same number of parameters (i.e., approximately 100,000 parameters) as PNN.

### III. EXPERIMENTAL RESULTS

#### A. Reduced Resolution Assessment

The reduced resolution assessment measures the similarity between a pansharpened HRMS image and a reference image (i.e., the original MS image). The similarity can be determined by several evaluation indices such as the spectral angle mapper (SAM) [68], the dimensionless global error in synthesis (ERGAS) [69], the spatial correlation coefficient (SCC) [70] and the Q2n [71] (e.g. Q8 for 8-band datasets and Q4 for 4-band datasets). The ideal values are 0 for SAM and ERGAS, and 1 for Q2n and SCC.

As described in Section II-F-1), there are two groups of test samples from WorldView-3 datasets (i.e., Tripoli and Rio). Each consisted of 1258 test samples and was processed for comparing SDRCNN with five representative CNN-based pansharpening methods (i.e., PNN, PanNet, BayesianNet, DMDNet and FusionNet). As shown in Table II, SDRCNN achieved the best average quantitative performance in all the metrics, demonstrating that it performed better than the other methods compared. This can be justified as compared to other CNN-based pansharpening methods, SDRCNN utilizes a dense residual connection mechanism to more fully exploit the residual information of different layers and achieves feature extraction for different levels from coarse to fine. In addition, combined with the proposed residual blocks, SDRCNN can obtain more detailed spatial features using a deeper network structure without increasing the number of parameters, thus improving the learning capability of the network.

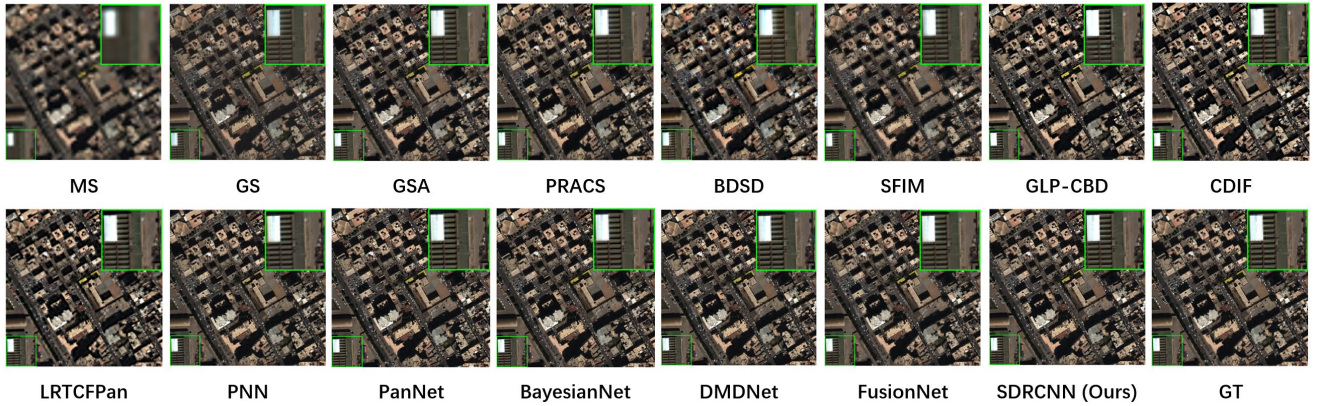

**Fig. 5.** Visual comparisons of the HRMS images from the compared methods on the reduced resolution Tripoli dataset (sensor: WorldView-3).

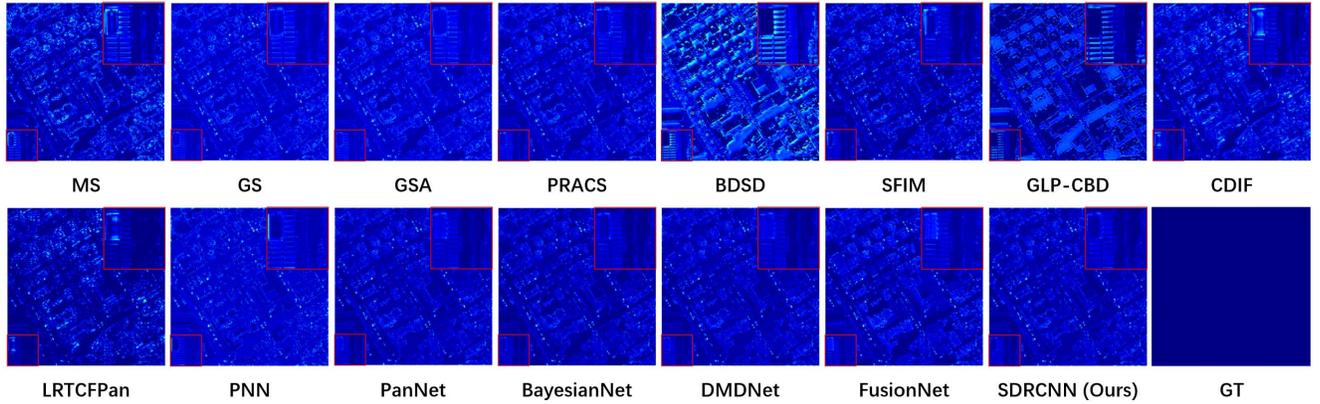

**Fig. 6.** Corresponding AEMs using the reference (GT) image on the reduced resolution Tripoli dataset (sensor: WorldView-3).

TABLE II
THE AVERAGE VALUES OF THE METRICS FOR PROPOSED SDRCNN AND ALL THE COMPARED CNNS, BASED ON 1258 REDUCED RESOLUTION SAMPLES:
(a) TRIPOLI DATASET AND (b) RIO DATASET.

| Method | (a) Tripoli dataset | | | | (b) Rio dataset | | | |
|---|---|---|---|---|---|---|---|---|
| | ERGAS (±std) | SAM (±std) | SCC (±std) | Q8 (±std) | ERGAS (±std) | SAM (±std) | SCC (±std) | Q8 (±std) |
| PNN | 4.6067±0.6705 | 6.4517±1.4093 | 0.9022±0.0402 | 0.8528±0.1306 | 3.0520±0.6450 | 3.9317±0.6641 | 0.8899±0.0580 | 0.8750±0.0860 |
| PanNet | 4.2245±0.6598 | 5.8269±1.1900 | 0.9115±0.0413 | 0.8674±0.1380 | 2.6781±0.4700 | 3.1122±0.5332 | 0.9367±0.0183 | 0.9272±0.0534 |
| BayesianNet | 4.1116±0.6402 | 5.7082±1.1886 | 0.9176±0.0391 | 0.8726±0.1318 | 2.5070±0.5210 | 2.9861±0.5368 | 0.9250±0.0154 | 0.9184±0.0351 |
| DMDNet | 4.1386±0.6417 | 5.7717±1.2069 | 0.9171±0.0381 | 0.8727±0.1314 | 2.5151±0.6464 | 3.0815±0.6339 | 0.9227±0.0529 | 0.9151±0.0698 |
| FusionNet | 4.2349±0.6372 | 5.8100±1.2270 | 0.9139±0.0384 | 0.8666±0.1323 | 2.5275±0.6313 | 2.9938±0.6174 | 0.9241±0.0521 | 0.9181±0.0690 |
| **SDRCNN** | **4.0567±0.5880** | **5.5678±1.1743** | **0.9211±0.0343** | **0.8765±0.1297** | **2.4980±0.5659** | **2.9763±0.6170** | **0.9269±0.0508** | **0.9190±0.0681** |
| Ideal value | 0 | 0 | 1 | 1 | 0 | 0 | 1 | 1 |

In addition, two test cases using Rio and Tripoli at a reduced resolution by applying the Wald's protocol stated in Section II-F-1) were generated. Fig. 5 and Fig. 7 show the visual comparisons of the pansharpened images derived using all the pansharpening methods considered for Tripoli and Rio, respectively. These visualisations were based on the three spectral channels Red, Green and Blue. For a better visualisation of their differences, a small local area (highlighted by the smaller square) was selected, zoomed in and shown in the larger square at the corner. Fig. 6 and Fig. 8 show the corresponding residuals, i.e., absolute error maps (AEMs), between the pansharpened images and GT images for Tripoli and Rio, respectively. The pansharpened images from the traditional methods (i.e., GS, GSA, PRACS, BDSD, SFIM, GLP-CBD, CDIF and LRTCFPan) showed some large spatial blurring and spectral distortions (indicated here by the colour distortions), particularly near the boundaries of buildings. All the CNN methods considered performed better (spatially and spectrally) than the traditional methods. Due to the size of the individual images shown in Fig. 5 and Fig. 7, it is not easy to distinguish the visual differences between the pansharpened images from the CNN methods. However, some differences can

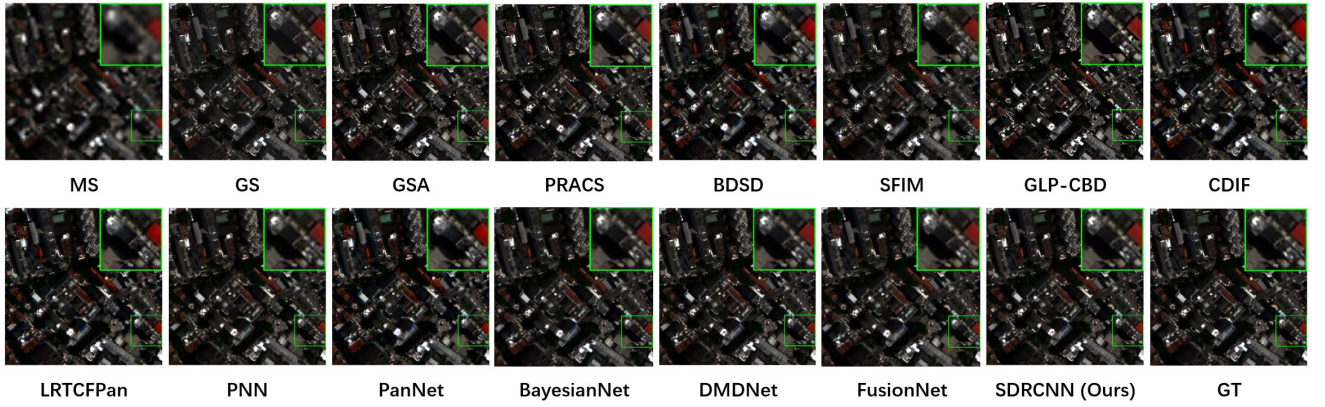

**Fig. 7.** Visual comparisons of the HRMS images from the compared methods on the reduced resolution Rio dataset (sensor: WorldView-3).

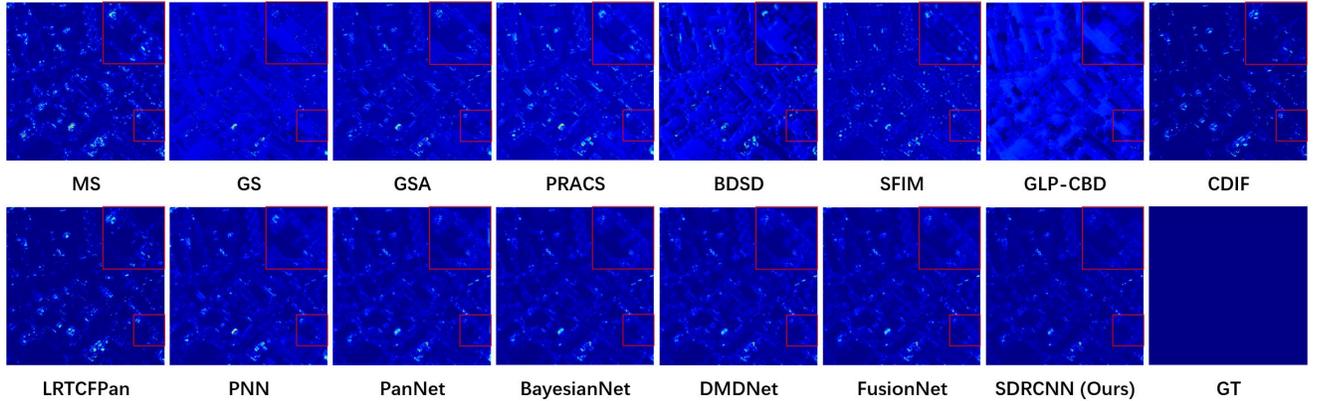

**Fig. 8.** Corresponding AEMs on the reduced resolution Rio dataset (sensor: WorldView-3).

TABLE III

QUALITY METRICS FOR ALL THE COMPARED APPROACHES ON THE REDUCED RESOLUTION TRIPOLI AND RIO DATASETS, RESPECTIVELY:
(a) TRIPOLI DATASET, (b) RIO DATASET.

| Method | (a) Tripoli dataset | | | | (b) Rio dataset | | | |
|---|---|---|---|---|---|---|---|---|
| | ERGAS | SAM | SCC | Q8 | ERGAS | SAM | SCC | Q8 |
| GS | 5.3307 | 6.9476 | 0.9310 | 0.8842 | 4.0774 | 4.4937 | 0.9451 | 0.9435 |
| GSA | 4.3501 | 6.1909 | 0.9352 | 0.9294 | 3.2516 | 3.9121 | 0.9462 | 0.9668 |
| PRACS | 4.4824 | 6.5096 | 0.9320 | 0.9228 | 3.3292 | 4.4638 | 0.9377 | 0.9668 |
| BDSD | 6.2745 | 8.5998 | 0.8795 | 0.8780 | 4.5744 | 5.9080 | 0.9367 | 0.9450 |
| SFIM | 4.9953 | 6.5628 | 0.9113 | 0.9076 | 3.5697 | 3.9780 | 0.9299 | 0.9606 |
| GLP-CBD | 4.8034 | 7.1610 | 0.9249 | 0.9308 | 5.1089 | 4.7943 | 0.9311 | 0.9367 |
| CDIF | 4.8266 | 6.8797 | 0.9341 | 0.9334 | 4.4216 | 4.0562 | 0.9516 | 0.9531 |
| LRTCFPan | 4.7632 | 6.6310 | 0.9350 | 0.9355 | 4.1438 | 3.9058 | 0.9558 | 0.9592 |
| PNN | 4.7168 | 6.9136 | 0.9285 | 0.9206 | 3.1055 | 4.2022 | 0.9443 | 0.9690 |
| PanNet | 4.1971 | 6.1086 | 0.9388 | 0.9374 | 2.6961 | 3.4805 | 0.9565 | 0.9759 |
| BayesianNet | 4.1126 | 6.0088 | 0.9434 | 0.9385 | 2.5537 | 3.3158 | 0.9601 | 0.9790 |
| DMDNet | 4.1425 | 6.0966 | 0.9424 | 0.9385 | 2.5582 | 3.4520 | 0.9600 | 0.9789 |
| FusionNet | 4.2815 | 6.1144 | 0.9380 | 0.9332 | 2.6283 | 3.3242 | 0.9595 | 0.9776 |
| **SDRCNN** | **4.0994** | **5.9123** | **0.9437** | **0.9404** | **2.5502** | **3.3057** | **0.9604** | **0.9796** |
| Ideal value | 0 | 0 | 1 | 1 | 0 | 0 | 1 | 1 |

be observed in Fig. 6 and Fig. 8. For example, in the AEMs of SDRCNN, the color of the zoomed region has a larger area tending to be the same dark blue as GT and showed less spectral residuals at the object boundaries, which suggests a good spectral preservation. Meanwhile, the AEMs of SDRCNN showed less details and textures than the other methods, which suggests that SDRCNN achieved the best spatial preservation. To better show their differences in performance, the values of the quantitative evaluation metrics were calculated and shown in Table III, which confirm that SDRCNN obtained the best performance with the smallest spatial and spectral distortions in the pansharpened images.

Fig. 9 and Fig. 10 show the loss curves with the iterations for the considered network structures trained on Tripoli and Rio datasets, respectively. To better evaluate the convergence of the networks, the vertical coordinates of the graphs use the mean value of the training loss for a total of 100 iterations before this iteration, thus reducing the fluctuation of the curves. It is clear that the SDRCNN exhibits the lowest test error after the curve has stabilized, which is consistent with the previous quantitative evaluation. In addition, the convergence rate of all these networks considered is approximately the same.

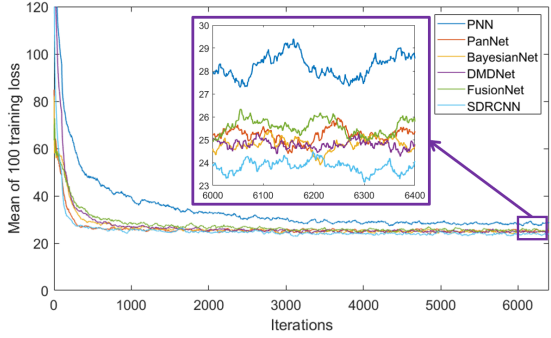

**Fig. 9.** Training loss curves for considered networks on Tripoli dataset (sensor: WorldView-3).

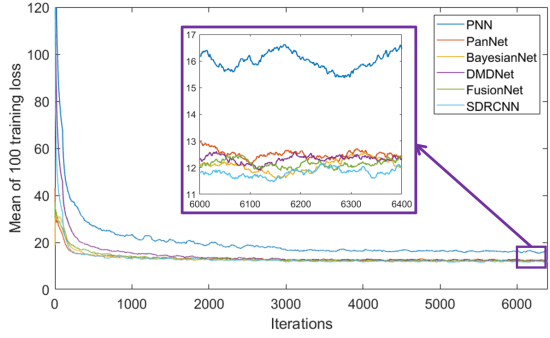

**Fig. 10.** Training loss curves for considered networks on Rio dataset (sensor: WorldView-3).

### B. Full Resolution Assessment

To confirm the results obtained at reduced resolutions, a full resolution analysis involving the original MS and PAN images was also needed. In this case, however, there are no GT images. For the full resolution assessments where GT images are absent, the widely used metrics in the literature include the quality with no reference (QNR) index, the spectral distortion $D_\lambda$ index, and the spatial distortion $D_s$ index [72], which were also adopted in our study. The ideal values of QNR, $D_\lambda$ and $D_s$ are 1, 0 and 0, respectively. In this case, the spectral quality of a fused image was referenced to the original MS image, while the spatial detail of a higher quality fused image should be more similar to that of the original PAN image. Table IV reports the average performances of 50 full resolution examples. On the indices considered, their values obtained for SDRCNN were closest to the ideal values with a small standard deviation, confirming the superiority of SDRCNN against the other methods compared.

### C. Assessment on WorldView-2 and QuickBird Datasets

To study the generalisation of SDRCNN, the proposed network was also tested using the data from another 8-band sensor (i.e., the Washington DC dataset acquired by WorldView-2) and a 4-band sensor (i.e., the QuickBird dataset). For a standard evaluation, the training and test data were generated according to the Wald's protocol. Visual comparisons of the results on the WorldView-2 dataset are shown in Fig. 11

| Method | QNR (±std) | $D_\lambda$ (±std) | $D_s$ (±std) |
|---|---|---|---|
| GS | 0.8601±0.0714 | 0.0505±0.0345 | 0.1073±0.0542 |
| GSA | 0.8829±0.0525 | 0.0461±0.0377 | 0.0948±0.0385 |
| PRACS | 0.9223±0.0392 | 0.0457±0.0178 | 0.0741±0.0272 |
| BDSD | 0.9116±0.0426 | 0.0443±0.0224 | 0.0824±0.0218 |
| SFIM | 0.8982±0.0347 | 0.0492±0.0368 | 0.0716±0.0277 |
| GLP-CBD | 0.8831±0.0559 | 0.0527±0.0357 | 0.0842±0.0340 |
| CDIF | 0.9271±0.0321 | 0.0430±0.0218 | 0.0501±0.0186 |
| LRTCFPan | 0.9273±0.0307 | 0.0428±0.0306 | 0.0498±0.0182 |
| PNN | 0.9267±0.0356 | 0.0428±0.0235 | 0.0514±0.0175 |
| PanNet | 0.9292±0.0364 | 0.0427±0.0230 | 0.0487±0.0201 |
| BayesianNet | 0.9335±0.0294 | 0.0421±0.0233 | 0.0444±0.0174 |
| DMDNet | 0.9326±0.0432 | 0.0422±0.0244 | 0.0445±0.0179 |
| FusionNet | 0.9308±0.0342 | 0.0426±0.0219 | 0.0461±0.0194 |
| **SDRCNN** | **0.9368±0.0274** | **0.0418±0.0212** | **0.0442±0.0168** |
| Ideal value | 1 | 0 | 0 |

and Fig. 12. It was observed that the fused image derived by SDRCNN was the most qualified one, evidenced not only by sharper and clearer edges, but also by unobservable ghosting and blurring. A visualisation of the output HRMS images on QuickBird was not provided as the difference could not be seen clearly with a naked eye. The values of the quality indices for the HRMS images obtained are reported in Table V for the two datasets, which shows that SDRCNN performed best amongst the methods considered. This suggests the generalisation of our approach.

### D. Visual Analysis of Dense Residual Connection

It was expected that the dense residual connection mechanism proposed in this study extracted different levels of features at different network depths and thus enhanced the spatial details of output images. To explore its effectiveness, the feature maps obtained at the three depths (i.e., three Addition Layers after Residual Blocks) in SDRCNN were visualized. For 52 feature maps obtained from each Addition Layer, it was considered as a data cube with 52 channels. Principal component analysis was performed using the singular value decomposition (SVD) method to achieve a dimensionality reduction of the data, resulting in four principal component bands. For visualization, the principal component values were rescaled to the range [0, 1]. Fig. 13 shows an example of a $256 \times 256$ size region taken from the Tripoli dataset of WorldView-3. The subplots in the first, second and third rows correspond to the visualised feature maps obtained from the Addition Layer after Residual Block 1, 2 and 3, respectively. It was observed that the feature maps in the first row included relatively coarse (i.e., larger objects) contour information. Although the maps in the second row showed similar edge information as those in the first row, these edges appeared to be darker and clearer, making the delineation between different objects or regions clearer. From the maps in the third row, more detailed (i.e., smaller objects) contour information and some building textures were observed. Overall, as the network depth increased, features extracted by SDRCNN varied from coarse to more detailed. In other words, the feature maps generated at a deeper level were found to have included more detailed spatial information. In order to better show the differences between the feature maps generated at the three depths in SDRCNN, Fig. 14 shows the AEMs obtained by subtracting the four principal component bands of the previous Addition Layer from those of the deeper Addition Layer. It was seen from the maps in the first row of Fig. 14 that the second Addition Layer included more spatial features than the first layer mainly in the edges of buildings. The maps associated with the third Addition Layer showed more detailed texture information of the objects/buildings than the second layer. These visual observations were reasonable and as expected. By connecting the residual information from the shallow layers and feeding it into the deeper convolutional layers for feature extraction, the residuals from different layers can be more fully utilized to obtain more detailed spatial features than the shallower layers of the network, thus improving the learning ability of the network for feature extraction.

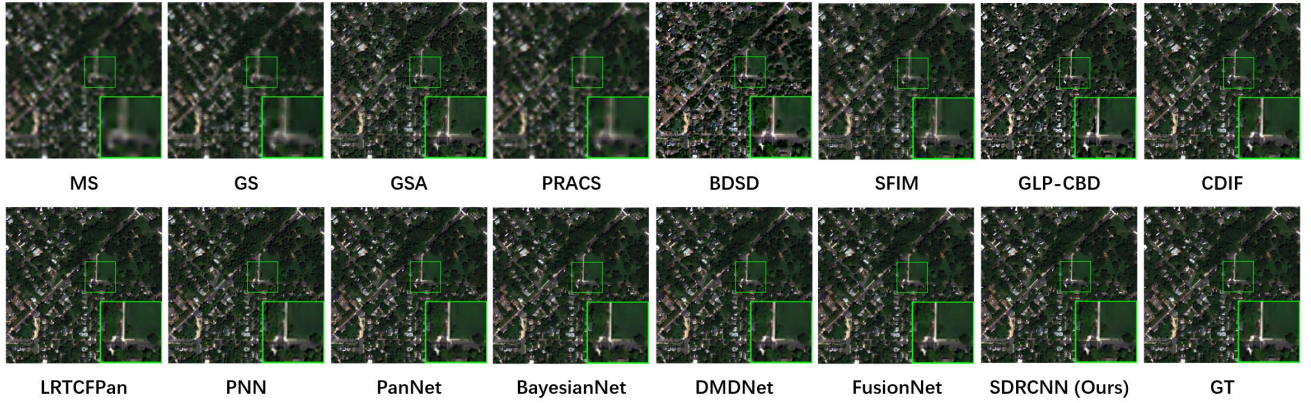

**Fig. 11.** Visual comparisons of the HRMS images from the compared methods on the reduced resolution Washington DC dataset (sensor: WorldView-2).

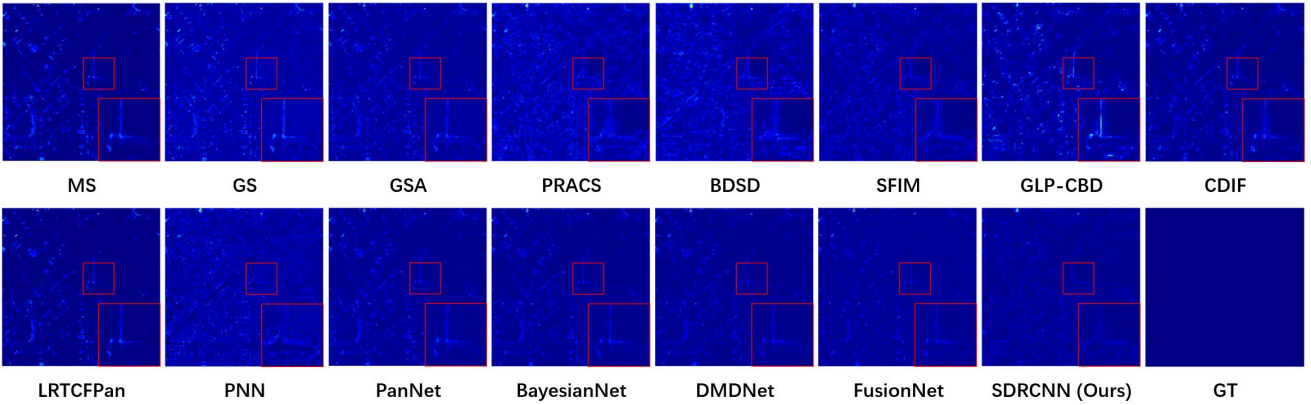

**Fig. 12.** Corresponding AEMs on the reduced resolution Washington DC dataset (sensor: WorldView-2).

TABLE V
QUALITY METRICS FOR ALL THE COMPARED APPROACHES ON THE REDUCED RESOLUTION WORLDVIEW-2 AND QUICKBIRD DATASETS, RESPECTIVELY:
(a) WORLDVIEW-2 (WASHINGTON DC) DATASET AND (b) QUICKBIRD DATASET.

| Method | (a) WorldView-2 (Washington DC) dataset | | | | (b) QuickBird dataset | | | |
|---|---|---|---|---|---|---|---|---|
| | ERGAS (±std) | SAM (±std) | SCC (±std) | Q8 (±std) | ERGAS (±std) | SAM (±std) | SCC (±std) | Q4 (±std) |
| GS | 4.5344±0.6049 | 6.3690±0.9200 | 0.8911±0.0420 | 0.8671±0.0727 | 1.6975±0.4013 | 2.2697±0.6203 | 0.9326±0.0200 | 0.7996±0.0611 |
| GSA | 3.6619±0.5000 | 6.0020±1.1338 | 0.8962±0.0341 | 0.9153±0.0715 | 1.4612±0.3640 | 1.9387±0.5798 | 0.9287±0.0229 | 0.8489±0.0720 |
| PRACS | 4.4238±0.5433 | 6.2247±1.1091 | 0.8516±0.0468 | 0.8770±0.0714 | 1.4128±0.3789 | 1.7971±0.5889 | 0.9313±0.0222 | 0.8449±0.0581 |
| BDSD | 5.0711±0.9168 | 8.3883±1.8437 | 0.8727±0.0373 | 0.8826±0.0722 | 1.7734±0.4468 | 2.3778±0.7730 | 0.9248±0.0193 | 0.8407±0.0554 |
| SFIM | 4.1077±2.4432 | 6.0334±1.0645 | 0.8839±0.0427 | 0.9014±0.0669 | 1.3934±0.3139 | 1.7945±0.5521 | 0.9319±0.0149 | 0.8474±0.0591 |
| GLP-CBD | 5.1367±0.7480 | 7.2313±1.2616 | 0.8468±0.0446 | 0.8895±0.0698 | 2.7821±0.6702 | 2.9029±0.9949 | 0.8922±0.0162 | 0.7647±0.0452 |
| CDIF | 3.7402±0.4158 | 6.0043±0.8574 | 0.9026±0.0116 | 0.9172±0.0711 | 1.6243±0.2547 | 1.9273±0.4177 | 0.9322±0.0105 | 0.8471±0.0620 |
| LRTCFPan | 3.7248±0.4389 | 5.7269±0.8634 | 0.9101±0.0230 | 0.9255±0.0726 | 1.5722±0.2566 | 1.8947±0.4203 | 0.9386±0.0142 | 0.8533±0.0623 |
| PNN | 3.6821±0.4414 | 6.1188±0.9339 | 0.8946±0.0239 | 0.9159±0.0730 | 1.5126±0.3558 | 1.9939±0.6281 | 0.9239±0.0184 | 0.8350±0.0533 |
| PanNet | 3.1710±0.3783 | 5.2083±0.8784 | 0.9304±0.0128 | 0.9330±0.0722 | 1.2476±0.2438 | 1.8423±0.4215 | 0.9412±0.0140 | 0.8766±0.0587 |
| BayesianNet | 2.9683±0.4076 | 4.8895±0.8464 | 0.9390±0.0095 | 0.9396±0.0713 | 1.0312±0.2416 | 1.3643±0.4021 | 0.9639±0.0096 | 0.8964±0.0610 |
| DMDNet | 3.0248±0.4132 | 4.9262±0.8720 | 0.9380±0.0102 | 0.9392±0.0706 | 1.0471±0.2464 | 1.4030±0.4195 | 0.9631±0.0102 | 0.8900±0.0607 |
| FusionNet | 3.0311±0.4229 | 4.9161±0.8587 | 0.9384±0.0097 | 0.9394±0.0719 | 1.0316±0.2414 | 1.3702±0.4058 | 0.9631±0.0102 | 0.8958±0.0612 |
| **SDRCNN** | **2.9563±0.4017** | **4.7633±0.8472** | **0.9404±0.0090** | **0.9404±0.0648** | **1.0309±0.2405** | **1.3514±0.3987** | **0.9649±0.0092** | **0.8970±0.0607** |
| Ideal value | 0 | 0 | 1 | 1 | 0 | 0 | 1 | 1 |

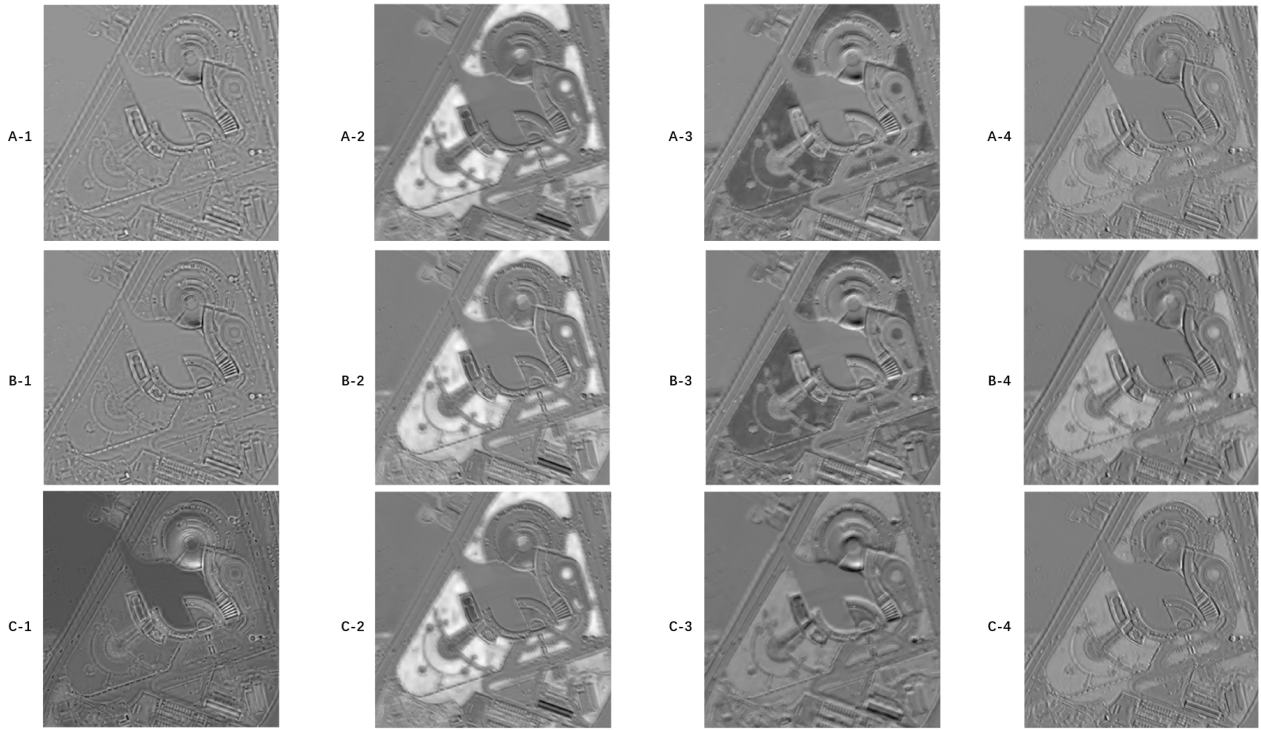

**Fig. 13.** Visual comparisons of the feature maps from three network depths of SDRCNN on the reduced resolution Tripoli dataset (sensor: WorldView-3). A, B and C correspond to the Addition Layers after the Residual Block 1, 2 and 3, respectively; the numbers 1, 2, 3 and 4 correspond to the four principal component bands generated by the dimensionality reduction of the data.

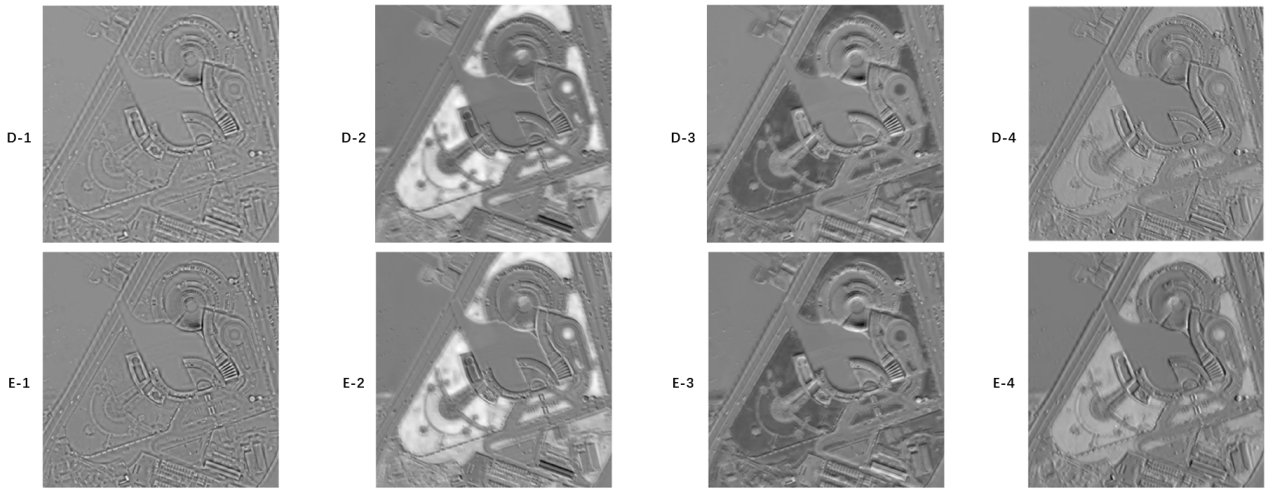

**Fig. 14.** Corresponding AEMs of the feature maps from two adjacent Addition Layers on the reduced resolution Tripoli dataset (sensor: WorldView-3). D corresponds to the residual of the shallower two Addition Layers, and E corresponds to the residual of the deeper two Addition Layers, the numbers 1, 2, 3 and 4 correspond to the four principal component bands generated by the dimensionality reduction of the data.

*E. Ablation Study*

Ablation studies were conducted to explore the effect of each component of SDRCNN. The WorldView-3 Tripoli dataset was used for these ablation studies.

*1) Network Structure of SDRCNN*

To verify the effectiveness of the single-branch single-scale network structure with a dense residual connectivity mechanism of SDRCNN, this ablation experiment replaced the structure of SDRCNN with those of other comparison networks (i.e., PNN, PanNet, BayesianNet, DMDNet and FusionNet) for comparison, respectively. Table VI shows the average results of the quantitative analysis of these combined networks and SDRCNN on 1258 test data from the Tripoli dataset. For a fair comparison, the number of parameters for all networks is 100K (K stands for $10^3$ ). The results show that the averaging performance of all metrics decreases when the network structure of SDRCNN was replaced with any other structure, thus demonstrating the effectiveness of the SDRCNN structure. It is worth noting that the quantification results corresponding to the network structures of PanNet and DMDNet in Table VI are the same, which is because of the same structures of these

TABLE VI

QUANTITATIVE COMPARISON OF THE NETWORKS COMBINING BASIC BLOCK OF SDRCNN AND **STRUCTURES** OF BENCHMARK METHODS (DL-BASED) ON 1258 TEST SAMPLES FOR TRIPOLI (WORLVIEW-3) DATASET

| **Structure** type of network | ERGAS (±std) | SAM (±std) | SCC (±std) | Q8 (±std) | Time (s) |
|---|---|---|---|---|---|
| PNN | 4.4544±0.6509 | 5.8912±1.2248 | 0.9095±0.0384 | 0.8615±0.1312 | 301.89 |
| PanNet | 4.1083±0.6057 | 5.6603±1.1909 | 0.9183±0.0349 | 0.8745±0.1315 | 338.14 |
| BayesianNet | 4.1292±0.6235 | 5.7045±1.1915 | 0.9173±0.0360 | 0.8740±0.1310 | 380.22 |
| DMDNet | 4.1083±0.6057 | 5.6603±1.1909 | 0.9183±0.0349 | 0.8745±0.1315 | 338.14 |
| FusionNet | 4.1031±0.5989 | 5.6822±1.2028 | 0.9187±0.0354 | 0.8740±0.1324 | 345.34 |
| **SDRCNN** | **4.0567±0.5880** | **5.5678±1.1743** | **0.9211±0.0343** | **0.8765±0.1297** | 327.46 |
| Ideal value | 0 | 0 | 1 | 1 | |

TABLE VII

QUANTITATIVE COMPARISON OF THE NETWORKS COMBINING STRUCTURE OF SDRCNN AND **BASIC BLOCKS** OF BENCHMARK METHODS (DL-BASED) ON 1258 TEST SAMPLES FOR TRIPOLI (WORLVIEW-3) DATASET

| **Block** type of network | ERGAS (±std) | SAM (±std) | SCC (±std) | Q8 (±std) | Time (s) |
|---|---|---|---|---|---|
| PNN | 4.2668±0.6933 | 5.8839±1.2231 | 0.9107±0.0423 | 0.8668±0.1313 | 349.54 |
| PanNet | 4.1978±0.6653 | 5.7783±1.1909 | 0.9124±0.0392 | 0.8724±0.1332 | 352.61 |
| BayesianNet | 4.2065±0.6674 | 5.7811±1.1996 | 0.9119±0.0411 | 0.8705±0.1329 | 417.49 |
| DMDNet | 4.1207±0.6276 | 5.7087±1.1937 | 0.9160±0.0378 | 0.8737±0.1311 | 793.61 |
| FusionNet | 4.2470±0.6439 | 5.8031±1.2143 | 0.9107±0.0386 | 0.8670±0.1323 | 355.99 |
| **SDRCNN** | **4.0567±0.5880** | **5.5678±1.1743** | **0.9211±0.0343** | **0.8765±0.1297** | 327.46 |
| Ideal value | 0 | 0 | 1 | 1 | |

TABLE VIII

QUALITY METRICS FOR ALL THE ABLATION STUDIES ON 50 TEST SAMPLES FOR TRIPOLI (WORLDVIEW-3) DATASET

| Method | ERGAS | SAM | SCC | Q8 | |
|---|---|---|---|---|---|
| PNN (without MS image connection in its original network) | 5.0635 | 7.0424 | 0.9152 | 0.9099 | |
| PNN with MS image connection | 4.7168 | 6.9136 | 0.9285 | 0.9206 | |
| SDRCNN without MS image connection | 4.1478 | 5.9468 | 0.9487 | 0.9396 | Section III-E-3) |
| **SDRCNN (with MS image connection by default)** | **4.0945** | **5.9078** | **0.9512** | **0.9432** | |
| SDRCNN with BN | 4.2110 | 6.0690 | 0.9388 | 0.9359 | |
| **SDRCNN (without BN by default)** | **4.0945** | **5.9078** | **0.9512** | **0.9432** | Section III-E-4) |
| SDRCNN with additional ReLU | 4.1369 | 5.9401 | 0.9410 | 0.9394 | |
| **SDRCNN (without additional ReLU by default)** | **4.0945** | **5.9078** | **0.9512** | **0.9432** | Section III-E-5) |
| Ideal value | 0 | 0 | 1 | 1 | |

TABLE IX

QUANTITATIVE COMPARISONS WITH THE BENCHMARK METHODS (PANNET, BAYESIANNET, DMDNET AND FUSIONNET) UNDER DIFFERENT PARAMETER NUMBERS ON 1258 TEST SAMPLES FOR TRIPOLI (WORLVIEW-3) DATASET

| Number of Parameters | Network | ERGAS (±std) | SAM (±std) | SCC (±std) | Q8 (±std) |
|---|---|---|---|---|---|
| | PanNet | 4.2444±0.6677 | 5.8561±1.1970 | 0.9104±0.0416 | 0.8658±0.1400 |
| | BayesianNet | 4.1758±0.6461 | 5.7610±1.2007 | 0.9141±0.0395 | 0.8695±0.1328 |
| ≈ 50K | DMDNet | 4.2375±0.6592 | 5.8354±1.2175 | 0.9103±0.0400 | 0.8697±0.1315 |
| | FusionNet | 4.2724±0.6352 | 5.8222±1.2279 | 0.9124±0.0386 | 0.8649±0.1325 |
| | **SDRCNN** | **4.1331±0.5998** | **5.6504±1.1967** | **0.9164±0.0348** | **0.8742±0.1306** |
| | PanNet | 4.2245±0.6598 | 5.8269±1.1900 | 0.9115±0.0413 | 0.8674±0.1380 |
| | BayesianNet | 4.1116±0.6402 | 5.7082±1.1886 | 0.9176±0.0391 | 0.8726±0.1318 |
| ≈ 100K | DMDNet | 4.1386±0.6417 | 5.7717±1.2069 | 0.9171±0.0381 | 0.8727±0.1314 |
| | FusionNet | 4.2349±0.6372 | 5.8100±1.2270 | 0.9139±0.0384 | 0.8666±0.1323 |
| | **SDRCNN** | **4.0567±0.5880** | **5.5678±1.1743** | **0.9211±0.0343** | **0.8765±0.1297** |
| | PanNet | 4.2025±0.6454 | 5.8012±1.1854 | 0.9126±0.0405 | 0.8679±0.1381 |
| | BayesianNet | 4.0869±0.6360 | 5.6513±1.1719 | 0.9184±0.0388 | 0.8732±0.1329 |
| ≈ 200K | DMDNet | 4.1091±0.6227 | 5.6791±1.1841 | 0.9180±0.0371 | 0.8738±0.1316 |
| | FusionNet | 4.1917±0.6273 | 5.7821±1.2134 | 0.9163±0.0385 | 0.8678±0.1325 |
| | **SDRCNN** | **4.0051±0.5893** | **5.4926±1.1530** | **0.9245±0.0335** | **0.8784±0.1307** |
| Ideal value | | 0 | 0 | 1 | 1 |

two networks. In addition, during this experiment, we found that the network framework of FusionNet is also the same as these two, but with a deeper structure than them by stringing one more basic block together. The last column of Table VI shows the corresponding testing times for each network. Since the PNN structure is much shallower than the others, it is reasonable that it has the shortest testing time. However, the testing times corresponding to the other structures are all longer than SDRCNN, thus further indicating that the SDRCNN structure is more computationally efficient.

### 2) Residual Block of SDRCNN

To demonstrate the effectiveness of the proposed residual block, this ablation experiment retains the network structure of the SDRCNN and replaces the blocks in the SDRCNN with the basic blocks of other comparison networks (i.e., PNN, PanNet, BayesianNet, DMDNet and FusionNet), respectively. Table VII shows the test results of these combined networks and SDRCNN (with the same 100K parameters) on the Tripoli dataset. It is obvious that the accuracy of all metrics decreases when the basic blocks in the SDRCNN are replaced with blocks from other networks, thus proving that better pansharpening accuracy can be obtained with blocks of SDRCNN. In addition, SDRCNN takes the shortest testing time, which further proves that the blocks of SDRCNN have higher efficiency.

### 3) Spectral Mapping

The spectral mapping refers to the operation of directly adding an upsampled MS image to an output image, and has been widely used in CNNs for pansharpening tasks. To verify whether it can improve the pansharpening accuracy of CNN, two networks, PNN and SDRCNN, were considered in this section for experiments. Table VIII shows the values of the accuracy metrics for PNN and SDRCNN with and without the spectral mapping on the 50 testdata. Compared with PNN, the fusion accuracy of the network with the spectral mapping were higher in all the metrics. In addition, the accuracy of pansharpening decreases when the spectral mapping of SDRCNN was removed. These suggest that the spectral mapping could enhance the performance of CNN for a higher pansharpening accuracy.

### 4) Removal of the BN Layer

In the existing CNNs for pansharpening, there is no agreement on whether the BN layer should be retained or not. To demonstrate the effectiveness of removing BN as described in Section II.D, experiments were conducted by plugging BN into our SDRCNN model. The BN operation was deployed after each convolution operation. Table VIII shows the average results of SDRCNN before and after the BN inclusion on 50 testdata using the WorldView-3 dataset. It was seen that the performance of SDRCNN with BN decreased. In addition, BN operations consumed additional computation and storage resources. As such, it is beneficial to exclude BN for higher pansharpening accuracy and improved computational and storage budget.

### 5) Reduced Activation Function

Similar to the problem of BN layer, there is no conclusion on the effect of activation function on the performance of CNN for pansharpening. In this section, the effectiveness of the reduced activation function (i.e., ReLU) described in section II.D was investigated by inserting more activation functions into our SDRCNN model. The additional activation functions were deployed after each depthwise convolution operation and concentration operation. Table VIII shows the average results of SDRCNN before and after the insertion of more activation functions on the 50 testdata. It was seen that better average results were obtained for the SDRCNN model retaining only one activation function between the two $1 \times 1$ layers.

### 6) Effect of the Number of Parameters

To check the effects of the network size on the performance of the networks (i.e., the benchmark methods PanNet, BayesianNet, DMDNet and FusionNet, and our method SDRCNN), three network sizes with approximately 50 K, 100K and 200K parameters were tested using the Tripoli dataset (containing 1258 test samples) from WorldView-3. Table IX shows the results of the quantity evaluation of the five CNN methods for the three network sizes considered. It can be seen that in each network size tested, SDRCNN achieved the closest average results to the ideal values and the smallest standard deviations in all quality metrics. In addition, as expected, increasing the network size can further improve the accuracy of SDRCNN.

## IV. DISCUSSION

Based on the experimental results of this study, the CNN-based methods exhibited better performances in pansharpening than the classical CS and MRA methods. This is mainly due to the fact that CNN-based methods exploit large-scale data during the training phase. In this section, the convolution kernel size, the network complexity and the testing time were discussed.

### 1) Convolution Kernel Size

Currently, some researchers argued that increasing the receptive field can improve the pansharpening performance of CNN models, and as such they used larger convolutional kernel sizes in their proposed methods [67]. This idea came from an observation of the behaviour of CNNs applied in some other image processing areas. In the proposed SDRCNN model, we also considered the behaviour of increasing the convolutional kernel. Initially, we tried using convolutional kernel sizes that are larger than 3×3, for example, 7×7 and 11×11. However, the reduction in network width or depth was necessary to control the network with a consistent number of parameters. In order not to change the width or depth of the network, we considered the use of the atrous convolution to explore the effect of increasing the receptive field. It was found that increasing the receptive field using the atrous convolution did not significantly improve the pansharpening performance of the CNN. It is possible that the limitations of the atrous convolution, such as grid effects, adversely affected the fusion results, which may lead to inaccurate results. In addition, it was also thought that a larger receptive field could improve the performance of CNNs in some other image processing domains, because these image processing tasks are usually highly dependent on the relationship between a pixel and its surrounding pixels, such as semantic segmentation and object detection, etc. However, for pansharpening, in the process of spatial and spectral information transfer and fusion, more attention should be paid to the relationship of features between images. In addition, larger convolution kernels usually bring a larger set of parameters, which often increase the computational cost and learning difficulty. Therefore, increasing the size of the convolution kernel or receptive field may not be suitable for improving the performance of the CNN model in the pansharpening task.

### 2) Network Complexity

SDRCNN is simpler than DMDNet, because SDRCNN does not need to compute high-pass filtered images from the input

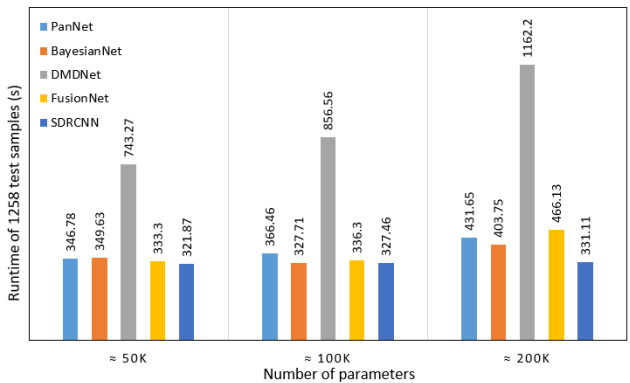

**Fig. 15.** Testing times for SDRCNN, DMDNet and FusionNet under different parameter numbers for 1258 test samples from Tripoli (WorldView-3) dataset.

PAN and MS images and uses smaller convolution kernels and a single-scale network structure. FusionNet replicates the input PAN image to the same number of bands as the input MS image and then feeds those into the network, which increases the workload of subsequent processing and feature extractions on the input images. In addition, SDRCNN has fewer layers than FusionNet. However, given that SDRCNN uses the depthwise separable convolutions, it is difficult to directly compare its network complexity with FusionNet. Furthermore, the structure of PNN is a simple three-layer network without any jump connections. However, it's too shallow to extract enough image features from this simple network. Therefore, SDRCNN not only improves the accuracy of pansharpening without increasing the network size, but also avoids unnecessary increases in network complexity. PanNet has a similar network structure and input data pre-processing requirement as DMDNet. BayesianNet is a multi-branch network, which is more complex than the single-branch single-scale SDRCNN.

### 3) Testing Time

Fig. 15 reports the testing times for PanNet, BayesianNet, DMDNet, FusionNet and SDRCNN on the Tripoli test set (containing 1258 test samples) from WorldView-3. DMDNet took more than twice as long as the other four networks, and the time spent on the testing increased notably with increasing number of parameters. As the number of parameters increases, SDRCNN always takes the least amount of time, and the efficiency of SDRCNN becomes more evident when the number of parameters reaches 200K.

## V. Conclusion

Developing highly accurate and efficient pansharpening methods can be very valuable. In this study, a novel single-branch, single-scale lightweight convolutional neural network architecture named SDRCNN was developed to this end. It uses a dense residual connected structure and novel convolution blocks to achieve a better trade-off between accuracy and efficiency. The performances of SDRCNN were compared to eight traditional methods (i.e., GS, GSA, PRACS, BDSD, SFIM, GLP-CBD, CDIF and LRTCFPan) and five lightweight CNN-based pansharpening methods (i.e., PNN, PanNet, BayesianNet, DMDNet and FusionNet), using the reduced-resolution and the full-resolution tests. SDRCNN achieved highest pansharpening accuracies (i.e., SAM, ERGAS, SCC, Q2n, QNR, $D_s$ and $D_\lambda$) on all four datasets tested. SDRCNN exhibited least spatial detail blurring and spectral distortions in visual comparisons using the pansharpened images and the associated AEMs. Furthermore, the ablation study confirmed the effectiveness of the dense residual connected structure and the convolutional block used in SDRCNN. Finally, it is demonstrated that SDRCNN is the most efficient among the compared lightweight networks. All these results demonstrated the superiority of SDRCNN against the traditional and lightweight CNN-based methods compared, in preserving both spatial and spectral information in fused HRMS images.

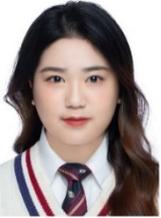

**Yuan Fang** received the B.E. degree in civil engineering from the University of Liverpool, UK, in 2021. She is now working toward her PhD degree at the same University.

Her research interests include deep learning and data fusion of satellite images.

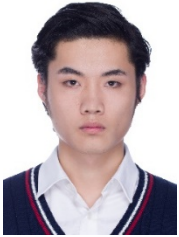

**Yuanzhi Cai** (Member, IEEE) received the B.E. degree in civil engineering from the University of Liverpool, UK, in 2020. He is currently pursuing the Ph.D. degree at the same University.

His research interests include the classification and segmentation of remote sensing data.

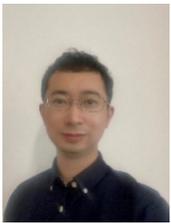

**Lei Fan** (Member, IEEE) received the Ph.D. degree from the University of Southampton, Southampton, UK, in 2018.

He is currently an Assistant Professor within Department of Civil Engineering at Xi'an Jiaotong Liverpool University, Suzhou, China. His main research interests include lidar and photogrammetry techniques, point cloud, machine learning, deformation monitoring, semantic segmentations, monitoring of civil engineering structures and geohazards.